\documentclass[final,authoryear,5p,times,twocolumn]{elsarticle}
\usepackage{amsmath}
\usepackage{amssymb} 

\usepackage{tikz}
\usetikzlibrary{arrows,matrix,positioning,patterns}
\usetikzlibrary{calc}
\usepackage{pgfplots} 
\usetikzlibrary{external}
\usetikzlibrary{angles}
\usepackage{hyperref}
\usetikzlibrary{plotmarks}

\newdefinition{rmk}{Remark}
\newproof{pf}{Proof}
\newproof{pot}{Proof of Theorem \ref{thm2}}

\usepackage[ruled, vlined, linesnumbered]{algorithm2e}

\usepackage{amsthm}

\newtheorem{proposition}{Proposition}

\newtheorem{problem}{Problem}

\usepackage{amsmath}
\usepackage{amssymb} 

\usepackage{eurosym}
\usepackage{colortbl}

\usetikzlibrary{arrows.meta,calc,decorations.markings,math,arrows.meta}

\usepackage[labelformat=simple]{subcaption}
\DeclareCaptionLabelSeparator{periodspace}{.\quad}
\usepackage{color,framed}
\usepackage[utf8]{inputenc} 
\usepackage{tabularx}

\usepackage{rotating}
\usepackage{graphicx}
\usepackage{lscape}
\allowdisplaybreaks 
\usepackage{longtable}

\usepackage[utf8]{inputenc}
\usepackage{pgfplots}
\usepgfplotslibrary{groupplots} 
\pgfplotsset{compat=1.3}

\makeatletter
\def\ps@pprintTitle{%
 \let\@oddhead\@empty
 \let\@evenhead\@empty
 \def\@oddfoot{}%
 \let\@evenfoot\@oddfoot}
\makeatother

\begin{document}

\begin{frontmatter}
%
%
\title{Path Planning for Spot Spraying with UAVs Combining TSP and Area Coverages}
\author{Mogens Plessen\corref{cor1}}
\cortext[cor1]{MP is with Findklein GmbH, Switzerland, \texttt{mgplessen@gmail.com}}

%
%
%
%

\begin{abstract}
This paper addresses the following task: given a set of patches or areas of varying sizes that are meant to be serviced within a bounding contour calculate a minimal length path plan for an unmanned aerial vehicle (UAV) such that the path additionally avoids given obstacles areas and does never leave the bounding contour. The application in mind is agricultural spot spraying, where the bounding contour represents the field contour and multiple patches represent multiple weed areas meant to be sprayed. Obstacle areas are ponds or tree islands. The proposed method combines a heuristic solution to a traveling salesman problem (TSP) with optimised area coverage path planning. Two TSP-initialisation and 4 TSP-refinement heuristics as well as two area coverage path planning methods are evaluated on three real-world experiments with three obstacle areas and 15, 19 and 197 patches, respectively. The unsuitability of a Boustrophedon-path for area coverage gap avoidance is discussed and inclusion of a headland path for area coverage is motivated. Two main findings are (i) the particular suitability of one TSP-refinement heuristic, and (ii) the unexpected high contribution of patches areas coverage pathlengths on total pathlength, highlighting the importance of optimised area coverage path planning for spot spraying.
\end{abstract}
\begin{keyword}
In-field path planning; Unmanned Aerial Vehicles; Traveling Salesman Problem; Area Coverage.
\end{keyword}
\end{frontmatter}


\section{Introduction\label{sec_intro}}

\begin{table}
\centering
\begin{tabular}{|ll|}
\hline
\multicolumn{2}{|c|}{MAIN NOMENCLATURE}\\
\multicolumn{2}{|l|}{Symbols}\\
$L$ & Pathlength, (m).\\
$N_\text{patches,all}$ & Total number of patches areas, (-).\\
$N_\text{patches,covg}$ & Nr. of patches needing area coverage logic, (-).\\
$T$ & Solution runtime, (s).\\
$W$ & Operating width (inter-lane distance), (m).\\
$c_{i,j}$ & Cost metric between two patches $i$ and $j$, (m).\\[3pt]
\multicolumn{2}{|l|}{Abbreviations}\\
DENN & Double-Ended Nearest Neighbour Algorithm.\\
NN & Nearest Neighbour Algorithm.\\
TSP & Traveling Salesman Problem.\\
UAV & Unmanned Aerial Vehicle.\\
\hline
\end{tabular}
\end{table}

\begin{figure}
\vspace{0.3cm}
\centering%
\begin{tikzpicture}
\draw[fill=white] (0,8) rectangle (8,9.4);
\node (c) at (4,8.7) [scale=1,color=black,align=left] {Data Input};
\draw[black,-{Latex[scale=1.0]}] (4, 8) -- (4, 7.4);
\draw[fill=white] (0,6) rectangle (8,7.4);
\node (c) at (4,6.7) [scale=1,color=black,align=left] {
Subprob. 1: TSP-path connecting patches areas
};
\draw[black,-{Latex[scale=1.0]}] (4, 6) -- (4, 5.4);
\draw[fill=white] (0,4) rectangle (8,5.4);
\node (c) at (3.62,4.7) [scale=1,color=black,align=left] {
For Each Patch:\\
 \hspace{0.55cm}Subprob. 2: Patch Area Coverage Path
};
\draw[black,-{Latex[scale=1.0]}] (4, 4) -- (4, 3.4);
\draw[fill=white] (0,2) rectangle (8,3.4);
\node (c) at (3.87,2.7) [scale=1,color=black,align=left] {
Connect one TSP-path and all coverage paths
};
%
\draw[black,-{Latex[scale=1.0]}] (4, 2) -- (4, 1.4);
\draw[fill=white] (0,0) rectangle (8,1.4);
\node (c) at (2.3,0.7) [scale=1,color=black,align=left] {Return path coordinates};
\end{tikzpicture}
\caption{High-level algorithm of this paper as a block diagram. A variety of heuristics are compared for Subproblem 1. Two main approaches are compared for Subproblem 2. Data input comprises one field contour, multiple patches areas contours, potential obstacles areas contours coordinates as well as a working width of the UAV.}
\label{fig_blockdiag}
\end{figure}

\begin{figure*}
\centering
\begin{subfigure}[t]{.33\textwidth}
  \centering
  \includegraphics[width=.99\linewidth]{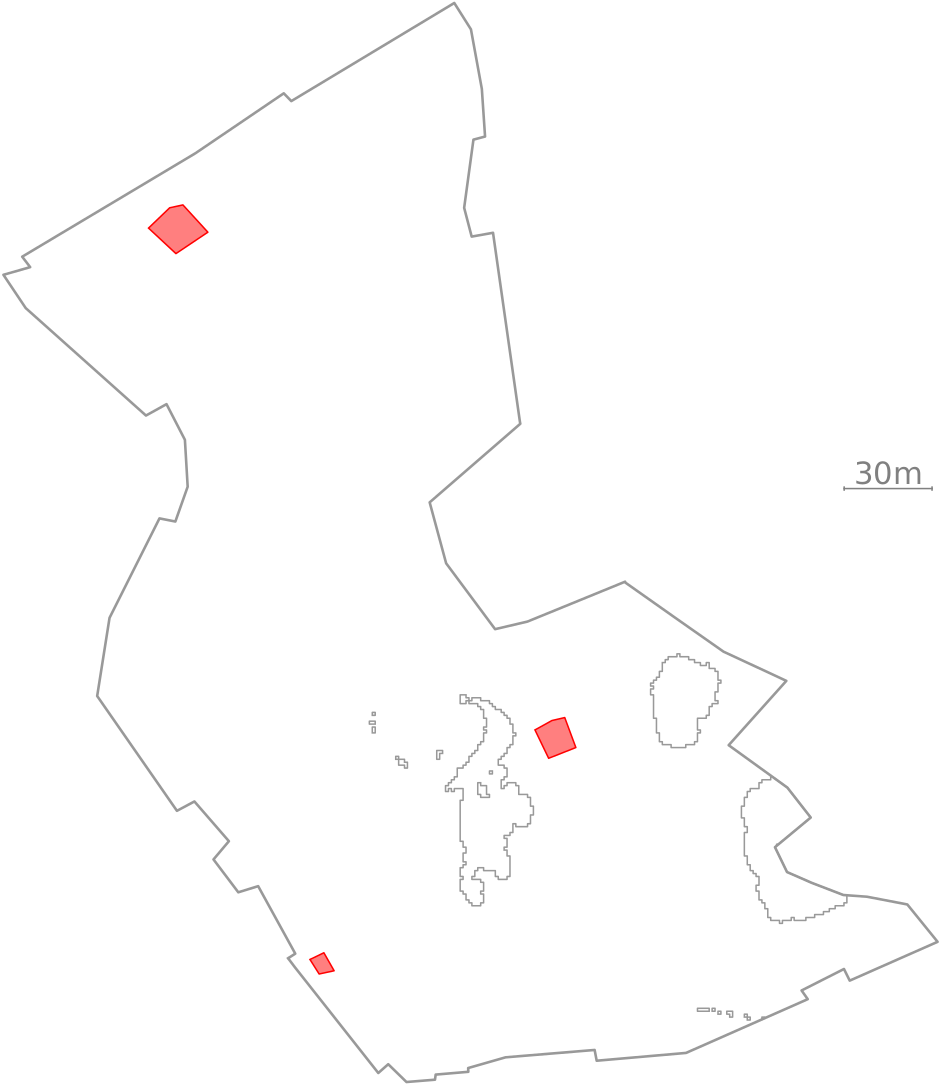}
\caption{Expt. 1: 15 patches}
  \label{fig_prob1}
\end{subfigure}%
\begin{subfigure}[t]{.33\textwidth}
  \centering
  \includegraphics[width=.99\linewidth]{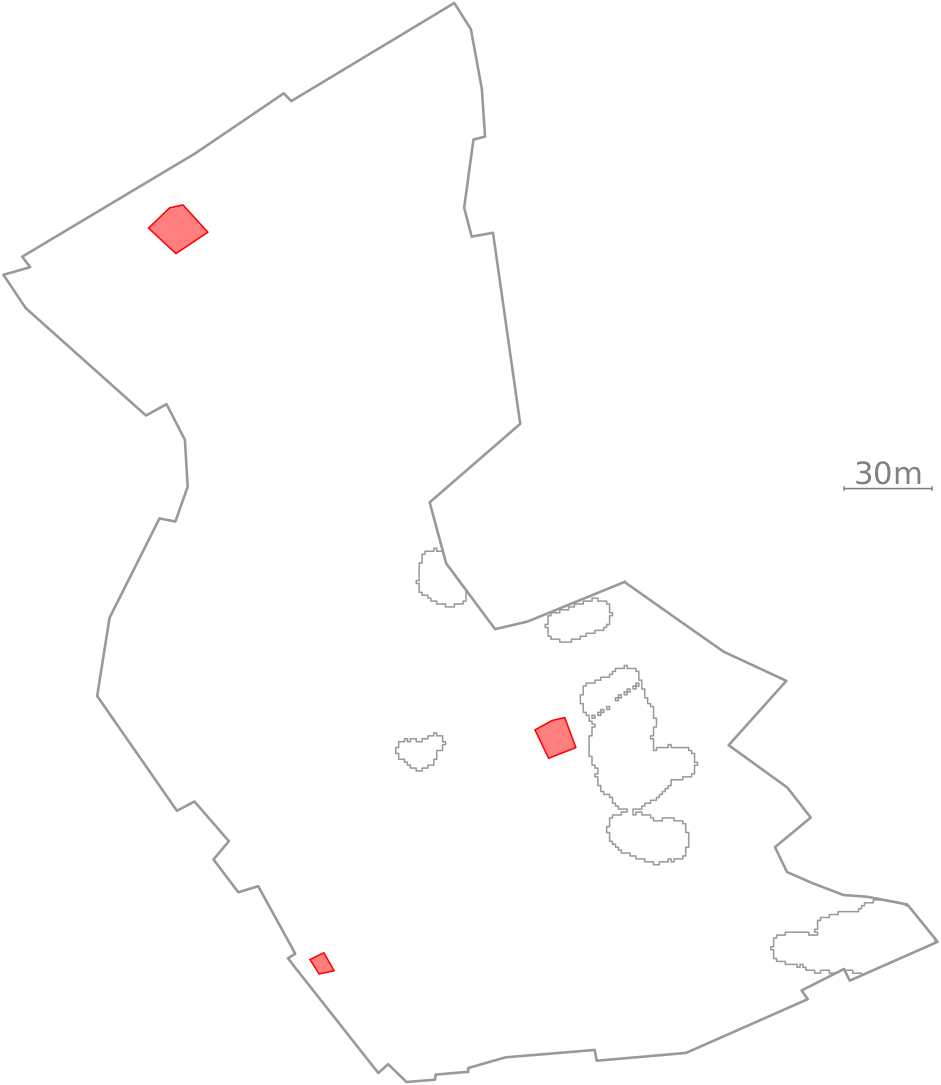}
\caption{Expt. 2: 19 patches}
  \label{fig_prob2}
\end{subfigure}
\begin{subfigure}[t]{.33\textwidth}
  \centering
  \includegraphics[width=.99\linewidth]{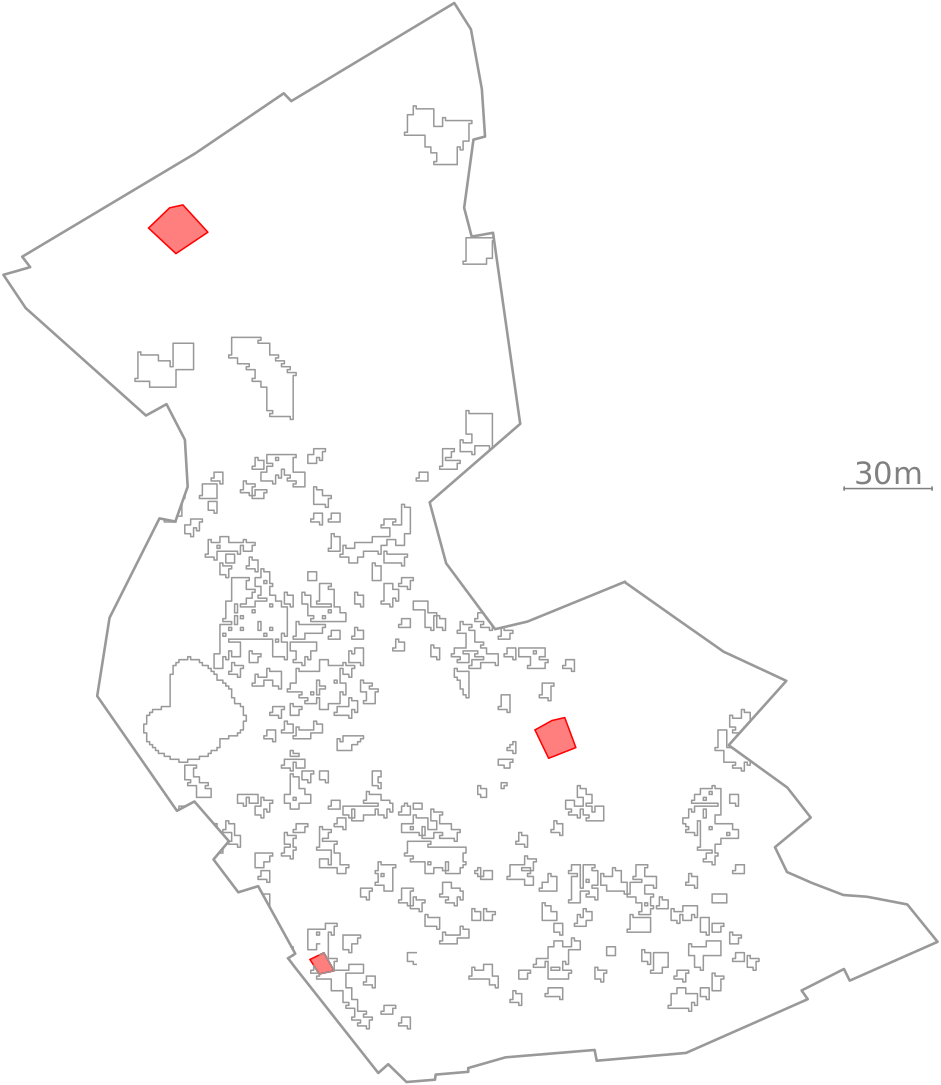}
\caption{Expt. 3: 197 patches}
  \label{fig_prob3}
\end{subfigure}
\caption{Problem visualisation: Given a set of patches or areas that are meant to be serviced within a field contour calculate a minimal length path plan for an UAV that additionally avoids any obstacles areas (red) and does not leave the field contour area. Three experiment setups of different complexity are displayed.}
\label{fig_probvisualisation}
\end{figure*}

The usage of unmanned aerial vehicles (UAVs) is growing in many engineering fields. Both for sensing as well as applying. Likewise, both for civilian (\cite{otto2018optimization}) as well as other applications (\cite{bortoff2000path}, \cite{kabamba2006optimal}). See \cite{kim2019unmanned} for a review specifically in agriculture.

One allure of UAVs is that, in contrast to ground-based vehicles that must account for roads or topographical constraints, for UAVs often direct line-of-sight transitions between two waypoints are enabled. A second major benefit is that subclasses of UAVs such as quadrotors permit near-holonomic paths, which is useful for applications requiring precision maneuvering.

Within the context of precision agriculture, the motivation and contribution of this paper is to present a path planning method for \emph{spot spraying} with UAVs: given a set of patches (e.g. weed or damaged areas, \cite{plessen2025accelerated}) of varying sizes that are meant to be serviced (with pesticide sprays) within a bounding contour (field contour) calculate a minimal length path plan for an UAV such that the path additionally avoids given obstacles areas (ponds and tree islands endangering the loss of the UAV) and never leaves the bounding contour (for safety reasons).

See \cite{hunter2020integration} for a related high-level description. For a review on spraying systems mounted on UAVs see \cite{hanif2022independent}. Here, the focus is on high-level path planning, specifically involving elements of a traveling salesman problem (TSP) for the transitions between patches areas and area coverage path planning for the coverage of these patches.

TSP is ubiquitous in logistics and extensively studied (\cite{bellmore1968traveling}). Nevertheless, it is a NP-hard problem and its solution is not obvious for different problem dimensions and applications. See \cite{furchi2022route} for a recent agricultural TSP-application. Likewise, area coverage methods greatly vary for different applications. One recurring method is based on the Boustrophedon-path (\cite{choset1998coverage}), e.g. see \cite{huang2023autonomous}, \cite{mukhamediev2023coverage}, \cite{li2023coverage} within the agricultural context. However, as will be shown, Boustrophedon-based area coverage is unisuitable  for precision spot spraying since area coverage gaps typically result. Remedy can be provided by inclusion of a headland path, which is typical for ground-based in-field path planning (\cite{plessen2019optimal}, \cite{plessen2018partial}), and here transfered to the UAV-setting.

In \cite{shah2020multidrone} and \cite{apostolidis2022cooperative} area coverage planning methods different from Boustrophedon-paths are used, but they also suffer from the absence of a headland path for area coverage gap avoidance. In \cite{deng2019constrained} the operation area is first partitioned into smaller subareas before mitered offset paths are used for area coverage. The disadvantage is that very tight heading changes for the subarea-innermost paths result, which in the limit can only be traced by a holonomic UAV.

A method combining both TSP and area coverage is presented in \cite{xie2018integrated}, where (i) Boustrophedon-based path planning is used for area coverage, (ii) at most 10 separate convexly-shaped patches are considered, and (iii) the TSP is solved by dynamic programming (DP). The authors acknowledge that the DP-approach is not scalable to larger dimensions. In a follow-up paper (\cite{xie2020path}) the same author group extends their method by heuristics to scale for up to 100 patches, however (i) maintaining a Boustrophedon-based area coverage logic, and (ii) assuming only convex polygonal area shapes. 

The research gap and motivation for this paper is discussed. For the application of spot spraying with UAVs a method combining both TSP with area coverage path planning is presented, whereby unlike in existing literature the inclusion of a headland path for each area coverage is explicitly assumed such that area coverage gaps can be minimised. The method is evaluated on a large-scale example with up to 197 arbitrarily non-convexly shaped patches areas.

The remaining paper is organised as follows: problem formulation, modeling and proposed solution, numerical results and the conclusion are described in Sections \ref{sec_probformulation}-\ref{sec_conclusion}.

\section{Problem Formulation\label{sec_probformulation}}

The problem addressed in this paper is visualised in Fig. \ref{fig_probvisualisation}.

\begin{problem}\label{problem1}
Given a set of patches of varying sizes that are meant to be serviced within a bounding contour calculate a minimal length path plan for an UAV such that the path additionally avoids given obstacles areas and does never leave the bounding contour.
\end{problem}

Geographic input data comprises coordinates of the field contour, obstacle areas contours and patches areas contours. UAV-related input data is here given by the UAV operating width, which is throughout this paper assumed as 2m. Throughout this paper it is assumed that all patches can be serviced by one flight. Partial patches servicing is left for future work. It is further assumed that the UAV begins and ends its path at the same location, the field entrance.

\section{Proposed Solution\label{sec_solnMethod}}

Any path plan for the given problem can be divided into two phases, (i) where the UAV travels between patches (no-spraying phase), and (ii) where the UAV covers a given patch (spraying phase). As will be shown, under specific assumptions these two problems can be treated separately without loss of optimality.

This section starts with problem modeling, before proposing solutions to both phases. 

\subsection{Problem Modeling and Subproblem Separation\label{sec_problmodeling}}

\begin{figure}
\centering
\begin{tikzpicture}
  \hspace*{-0mm}\node at (0,0) [opacity=1.0] {\includegraphics[width=.8\linewidth,keepaspectratio]{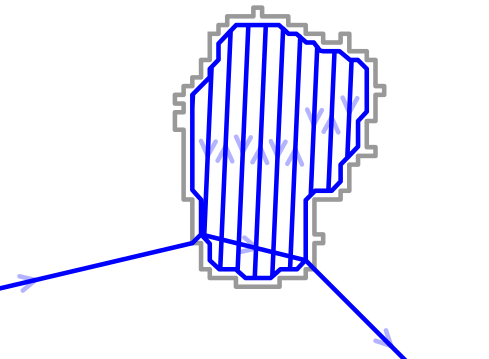}};  
  \node (c) at (-1.7,-0.7) [scale=1.0,color=black,align=left] {patch entry\\[-4pt]point};
  \node (c) at (2.0,-1.15) [scale=1.0,color=black,align=left] {patch exit\\[-4pt]point};
  \node (c) at (3.4,-2.4) [scale=1.0,color=black,align=left] {towards next\\[-4pt]patch};	  
\end{tikzpicture}
\caption{Illustration of patch entry and exit point, and an exemplary patch area coverage path plan including a headland path locally in parallel to the patch contour (gray).}
\label{fig_patchentryexit}
\end{figure}

Let a transition graph model costs $c_{i,j}>0$ for the transition between any two patches $i$ and $j$ for all $i,j\in\{1,\dots,N_\text{patches,all}\}$, with $N_\text{patches,all}>0$ denoting the total number of patches. Costs can be modeled in a variety of ways. Three ways are mentioned.

First (and used throughout the remainder of this paper), $c_{i,j}$ can denote the shortest pathlength between any two patches $i$ and $j$ by connecting the closest projection points between any two patches. These projection points later serve as patch entry and exit points for any TSP-path covering all patches. For visualisation see Fig. \ref{fig_patchentryexit}.

Second, centroids of patches could be used to calculate transition costs by intersecting lines between centroids with their corresponding patch contours to generate patch entry and exit points.

Third, instead of pathlength a different heuristic cost metric (typically a weighted pathlength) might be used that renders the problem non-symmetric, i.e., $c_{i,j}\neq c_{j,i}$. Especially in open-space path planning this is merited, where wind might be taken into account. 

\begin{figure}
\centering
\includegraphics[width=.3\linewidth]{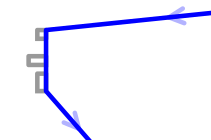}
\caption{Illustration of a TSP-segment covering three small patches (gray) that do not require an area coverage path plan for a given larger UAV operating width.}
\label{fig_smallpatches}
\end{figure}

In general, patches areas are of varying size and arbitrarily non-convexly shaped. For small patches and a given UAV-operating width a flight from one projection point (the patch entry point) on towards another projection point (the patch exit point) can be sufficient to fully spray that patch, see Fig. \ref{fig_smallpatches}. In contrast, for larger patches an area coverage path planning logic is required, see Fig. \ref{fig_patchentryexit}. The number of patches requiring such logic shall be denoted by $N_\text{patches,covg}\geq 0$.

\begin{proposition}\label{def_proposition1}
For a given transition graph connecting all patches and thereby naturally also fixing patch entry and exit points for any path along that transition graph, and under the assumption of (i) permitted straight flight lines between patch entry point and patch exit point, and (ii) area coverage path plans that must include a headland path (i.e., no pure Boustrophedon paths permitted), Problem \ref{problem1} can without loss of optimality be separated into two subproblems, one solving a TSP for connecting the patches, and one for separately calculating an area coverage path plan for each patch area.
\end{proposition}
\begin{pf}
For a given transition graph, the exact TSP-solution by definition finds a shortest path covering all patches. This TSP-solution automatically gives rise to a sequence of patch entry and exit points. The shortest possible path between any patch entry and exit point is a straight line.
What remains to be shown is that an area coverage path for a given patch starting and ending at the patch entry point plus then transitioning along a straight line towards the patch exit point is the optimal solution. Under the assumption that the area coverage path plan must include a headland path this is indeed the case by construction of an Eulerian cycle according to the method in \cite{plessen2019optimal}. Finding such an area coverage path plan is independent from the remainder of the TSP-solution and therefore proves the separability property.
\end{pf}

Several comments are made. First, the key assumption that enables the separability property is that transition graph generation with patch entry and exit points (the choice of the aforementioned three cost metrics does not matter) must be \emph{fixed before} calculating the area coverage path plans. Conversely, if freely permitting to optimise patch entry and exit points as a function of area coverage pathlength, TSP-solution and area coverage path planning for the patches become interrelated, and a trade-off between specific patch entry and exit points suitable for area coverage in combination with their embedding in a TSP-solution might then be optimal. In such a scenario, area coverage pathlength savings derived from optimized patch entry points might overall be larger than savings from a TSP-solution for the transition graph generated via shortest distances between the projection points of all patches.

Second, likewise (by providing a similar argument) optimality while maintaining the separability property is in general violated when after area coverage, instead of straight lines between patch entry and exit point, alternative paths such as e.g. a path along the headland (for example motivated by an UAV with a leaking tank wishing to avoid leaking into the central interior patch area), between patch entry and exit point are used.

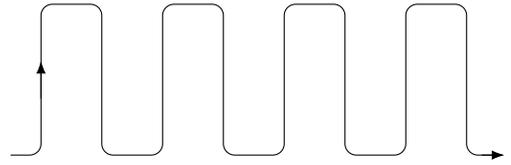
\begin{figure}
\vspace{0.3cm}
\centering%
\begin{tikzpicture}
\draw [black] plot [rounded corners=0.15cm] coordinates { (0.4,0)(0.8,0)(0.8,2)(1.6,2)(1.6,0)
(2.4,0)(2.4,2)(3.2,2)(3.2,0)
(4.0,0)(4.0,2)(4.8,2)(4.8,0)
(5.6,0)(5.6,2)(6.4,2)(6.4,0)
(6.8,0)
};
\draw [black,-{Latex[scale=1.0]}] plot [rounded corners=0.25cm] coordinates { (6.6,0)(6.9,0) };
\draw [black,-{Latex[scale=1.0]}] plot [rounded corners=0.25cm] coordinates { (0.8,0.75)(0.8,1.25) };
\end{tikzpicture}
\caption{Illustration of a Boustrophedon path. Characteristic is the zigzag-like path and the absence of a headland path.}
\label{fig_Boustrophedon}
\end{figure}

Third, arguments why Boustrophedon paths (see Fig. \ref{fig_Boustrophedon}) are deemed unsuitable in this paper are given below in Section \ref{subsec_patchesareacovg} when discussing area coverage path planning in detail.

Under the assumptions and according to Proposition \ref{def_proposition1} the TSP-solution and patches area coverage are treated as two separate problems within the framework of Fig. \ref{fig_blockdiag}. The two subproblems are discussed in the following two subsections.

\subsection{Subproblem 1: TSP}


The basic TSP-problem is: given $N$ locations find a minimum-length tour that visits each location exactly once (\cite{bellmore1968traveling}). Finding an optimal TSP-solution is a NP-hard problem. For $N+1$ locations ($N$ patches plus the field entrance location) there are $N\cdot(N-1)\cdot(N-2)\cdot\dots\cdot 1$ unique TSP-paths. In general, the TSP-problem can be cast into a dynamic programming framework (see \cite{bertsekas2012dynamic}) and solved by a label correcting method similarly as in \cite{plessen2016shortest}. However, for Expt. 3 in Fig. \ref{fig_prob3} with 197 patches unimaginable $\mathcal{O}(10^{368})$ unique TSP-combinations result. Thus, for problem dimensions at hand a heuristic solution \emph{must} be devised. 

In the following, let $S=(s[0],s[1],\dots,s[N_\text{patches,all}],s[0])$ represent a TSP-path that starts and ends at the field entrance location and visits each of $N_\text{patches,all}$ patches exactly once. In line with TSP-literature, we use the terms 'location', 'patch' or 'node' interchangeably. The symbol $N_\text{patches,all}$ instead of a more generic $N$ is used since $N_\text{patches,covg}\leq N_\text{patches,all}$ of the larger patches require additional area coverage path planning.

In this paper, two TSP-initialisation and 4 TSP-refinement heuristics are discussed. For final experiments, also 2 combinations of the TSP-refinement heuristics are additionally evaluated. The focus is on simple and hyperparameter-free heuristics, which excludes e.g. space partitioning-based heuristics (\cite{huang2017investigating}). 

The two TSP-initialisation heuristics, that build a first TSP-path from scratch, are:
\begin{itemize}
\item[\textsf{NN}] \emph{Nearest neighbour}. A classic heuristic (\cite{bellmore1968traveling}). Starting from the field entrance, the closest patch according to the transition graph cost and not yet part of the TSP-solution is iteratively added to the TSP-path until all patches are included.
\item[\textsf{DENN}] \emph{Double-ended nearest neighbour}. Inspired by \cite{huang2017investigating} \emph{two} NN-branches are simultaneously build starting from the field entrance location. In the last iteration both branches are connected to generate a TSP-path. 
\end{itemize}

Note that both methods are deterministic. The four TSP-refinement heuristics considered refining a given TSP-path are:
\begin{itemize}
\item[\textsf{H$_1$}] \emph{Pairwise random switching}. Draw two random indices $1\leq i,j\leq N_\text{patches,all}$, switch their position within the TSP-path, and if this switching yields a lower-cost TSP-path then keep this modification as updated TSP-path. 
\item[\textsf{H$_2$}] \emph{Remove a node and insert it somewhere}. Draw a random index  $1\leq i\leq N_\text{patches,all}$, remove the corresponding node and insert it at another randomly drawn index $1\leq j\leq N_\text{patches,all}$, and if this yields a lower-cost TSP-path then keep this modification as updated TSP-path.
\item[\textsf{H$_3$}] \emph{Reversion}. Draw a random index $1\leq i\leq N_\text{patches,all}$, flip the two consecutive nodes starting at that index such that $S=(s[0],\dots,s[i+1],s[i],\dots,s[0])$ results, and if this yields a lower-cost TSP-path then keep this modification as updated TSP-path.
\item[\textsf{H$_4$}] \emph{Encourage no crossings}. For-Loop consecutively through all indices $\{i:0\leq i\leq N_\text{patches,all}-1\}$ of the TSP-path, at every index $i$ select the node at that index and the immediately next node, i.e., $s[i]$ and $s[i+1]$. Embeddedly, while-loop through all indices $\{j:i+1\leq j\leq N_\text{patches,all}-1\}$ of the TSP-path, at every index $j$ select the node at that index and the immediately next node, i.e. $s[j]$ and $s[j+1]$. Check if the two lines constructed by $(s[i],s[i+1])$ and $(s[j],s[j+1])$ are not in parallel. If they are not in parallel, try to reroute those 4 nodes such that a lower-cost TSP-path results. If this rerouting yields a lower-cost TSP-path then keep this modification as updated TSP-path, break the while-loop, and increment $i+1$ in the outer for-loop. Continue executing and iterating for-loops until no further cost improvements are achieved, i.e., until one for-loop run does not yield any TSP-patch update.
\end{itemize}

Several comments are made. First, note that \textsf{H$_1$}, \textsf{H$_2$} and \textsf{H$_3$} are random sampling-based heuristics. A desired walltime limit is set (e.g. 10s) and these heuristics are executed for that time. The allure of this type of heuristics is that they are simple and suitable for parallelisation. For example, \textsf{H$_2$} is also used in \cite{plessen2020gpu} for GPU-accelerated logistics optimisation for biomass production. As show in Section \ref{sec_IllustrativeEx}, out of the three random sampling-based heuristics \textsf{H$_2$} also performed best in this paper. Second and in contrast, \textsf{H$_4$} is a deterministic heuristic. In practice, it was found to converge very quickly and, as will be shown, turned out to be the most useful TSP-refinement heuristic of the four options. The intuitive interpretation of \textsf{H$_4$} is that TSP-paths are encouraged that reduce crossings and edgy paths. Third, for \textsf{H$_4$} it was also tested to conduct the rerouting step only when the \emph{line segments} constructed by $(s[i],s[i+1])$ and $(s[j],s[j+1])$ were really intersecting (instead of just not being parallel). It was empirically found that the relaxed criterion of not being parallel was much more efficient in generating TSP-path improvements.

\subsection{Subproblem 2: Patches Area Coverage Path Planning\label{subsec_patchesareacovg}}

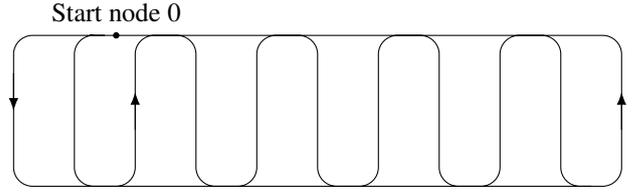
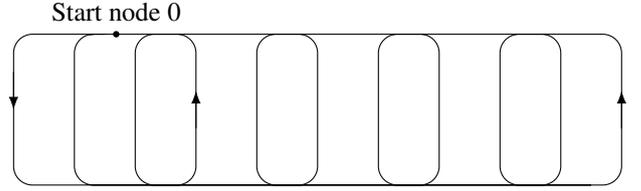
\begin{figure}
\vspace{0.3cm}
\begin{subfigure}[b]{\linewidth}
\centering%
\begin{tikzpicture}
\draw [black] plot [rounded corners=0.25cm] coordinates { (0.4,2)(-0.8,2)(-0.8,0)(7.2,0)(7.2,2)(0.45,2)
(0,2)(0,0)(0.8,0)(0.8,2)(1.6,2)(1.6,0)
(2.4,0)(2.4,2)(3.2,2)(3.2,0)
(4.0,0)(4.0,2)(4.8,2)(4.8,0)
(5.6,0)(5.6,2)(6.4,2)(6.4,0)
(6.8,0)
};
\draw[fill=black] (0.55,2) circle (1pt); 
\node[color=black] (a) at (0.55, 2.3) {Start node 0};
\draw [black,-{Latex[scale=1.0]}] plot [rounded corners=0.25cm] coordinates { (-0.8,1.5)(-0.8,1)};
\draw [black,-{Latex[scale=1.0]}] plot [rounded corners=0.25cm] coordinates { (7.2,0.75)(7.2,1.25) };
\draw [black,-{Latex[scale=1.0]}] plot [rounded corners=0.25cm] coordinates { (0.8,0.75)(0.8,1.25) };
\end{tikzpicture}
\caption{\textsf{Classic} / Boustrophedon with headland path.}
\end{subfigure}\\%
\begin{subfigure}[b]{\linewidth}
\centering%
\begin{tikzpicture}
\draw [black] plot [rounded corners=0.25cm] coordinates { (0.4,-3.25)(-0.8,-3.25)(-0.8,-5.25)(7.2,-5.25)(7.2,-3.25)(0.45,-3.25)(0,-3.25)
(0,-5.25)(1.6,-5.25)(1.6,-3.25)(0.8,-3.25)(0.8,-5.25)
(1.6,-5.25)(3.2,-5.25)(3.2,-3.25)(2.4,-3.25)(2.4,-5.25)
(3.2,-5.25)(4.8,-5.25)(4.8,-3.25)(4,-3.25)(4,-5.25)
(4.8,-5.25)(6.4,-5.25)(6.4,-3.25)(5.6,-3.25)(5.6,-5.25)
(6.4,-5.25)(6.8,-5.25)
};
\draw[fill=black] (0.55,-3.25) circle (1pt); 
\node[color=black] (a) at (0.55, -2.95) {Start node 0};
%
\draw [black,-{Latex[scale=1.0]}] plot [rounded corners=0.25cm] coordinates { (-0.8,-3.75)(-0.8,-4.25) };
\draw [black,-{Latex[scale=1.0]}] plot [rounded corners=0.25cm] coordinates { (7.2,-4.5)(7.2,-4.0) };
\draw [black,-{Latex[scale=1.0]}] plot [rounded corners=0.25cm] coordinates { (1.6,-4.5)(1.6,-4.0) };
\end{tikzpicture}
\caption{\textsf{Optimised} / naturally resulting pattern emphasised in \cite{plessen2018partial}.}
\end{subfigure}
\caption{Sketch of two path planning patterns for coverage of a rectangular-shaped exemplary area.}
\label{fig_sketch_2pathplanningmethods}
\end{figure}

\begin{figure}
\centering
\begin{subfigure}[t]{.5\textwidth}
  \centering
  \begin{tikzpicture}
  \hspace*{-0mm}\node at (0,0) [opacity=1.0] {\includegraphics[width=.8\linewidth,keepaspectratio]{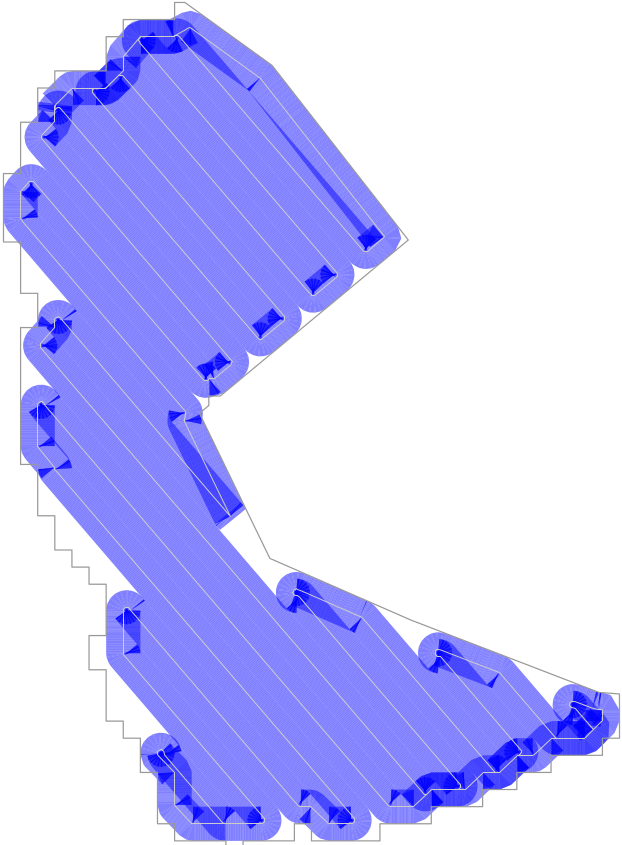}};  
  
  \draw[red,ultra thick] (0.,1.3) circle [x radius=1.3cm, y radius=5mm, rotate=40]; 
  \node (c) at (0,0) [scale=1.5,color=red,align=left] {Type 1};
  
  \draw[red,ultra thick] (-2.8,-1.35) circle [x radius=1.3cm, y radius=5mm, rotate=130];
  \node (c) at (-3.7,-2.2) [scale=1.5,color=red,align=left] {Type 2};
  
  \draw[red,ultra thick] (-3.3,1.3) circle [x radius=1.3cm, y radius=5mm, rotate=90];
  \draw[red,ultra thick] (-2.1,-3.5) circle [x radius=1.3cm, y radius=5mm, rotate=130];
  \draw[red,ultra thick] (-0.6,-1.5) circle [x radius=1.3cm, y radius=5mm, rotate=130];
  \draw[red,ultra thick] (-0.6,-1.5) circle [x radius=1.3cm, y radius=5mm, rotate=130];
  \draw[red,ultra thick] (1.2,-2.5) circle [x radius=0.8cm, y radius=0.5cm, rotate=150];
  \draw[red,ultra thick] (2.8,-3.1) circle [x radius=0.7cm, y radius=0.5cm, rotate=150];
\end{tikzpicture}
\caption{Boustrophedon path: characteristically incurring area coverage gaps. (Type 1) Lanes are circa perpendicular to the area contour. (Type 2) More profound area coverage gaps occur when lanes are almost aligned with the area contour.}
  \label{fig_Boustrophedon_areacovg}
\end{subfigure}
\begin{subfigure}[t]{.5\textwidth}
  \centering
  \includegraphics[width=.8\linewidth]{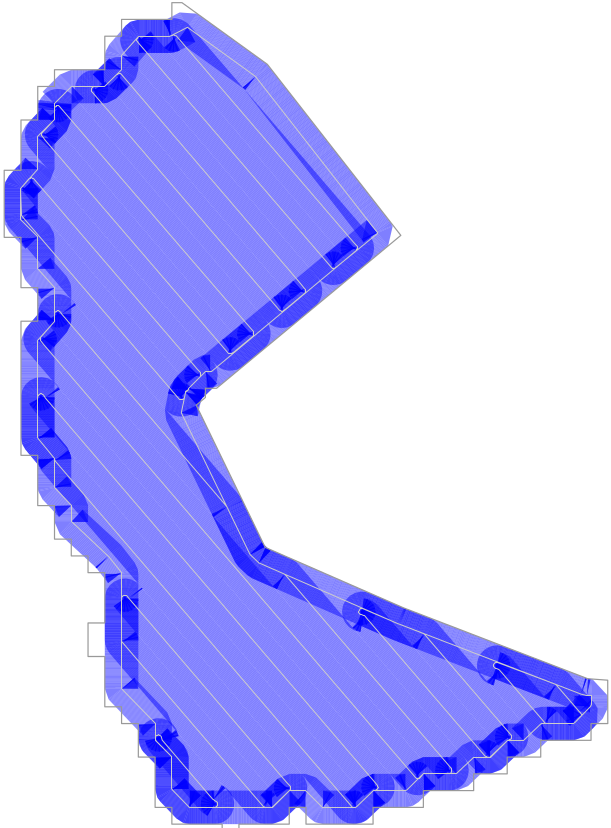}
\caption{Minimised area coverage when also including a headland path.}
  \label{fig_incldheadl_areacovg}
\end{subfigure}
\caption{Illustration of area coverage and overlapping area. Notice how for Boustrophedon-based paths significant area coverage gaps occur. This is not acceptable for precision spot spraying. (As extending detail, it is here assumed that UAV-spraying is always switched on, which is indicated by the darker blue overlapping areas. In practice, overlap areas can be reduced by appropriate boom section control and individual spray nozzle control.)}
\label{fig_areacovg}
\end{figure}

\begin{figure}
\centering
\includegraphics[width=.7\linewidth]{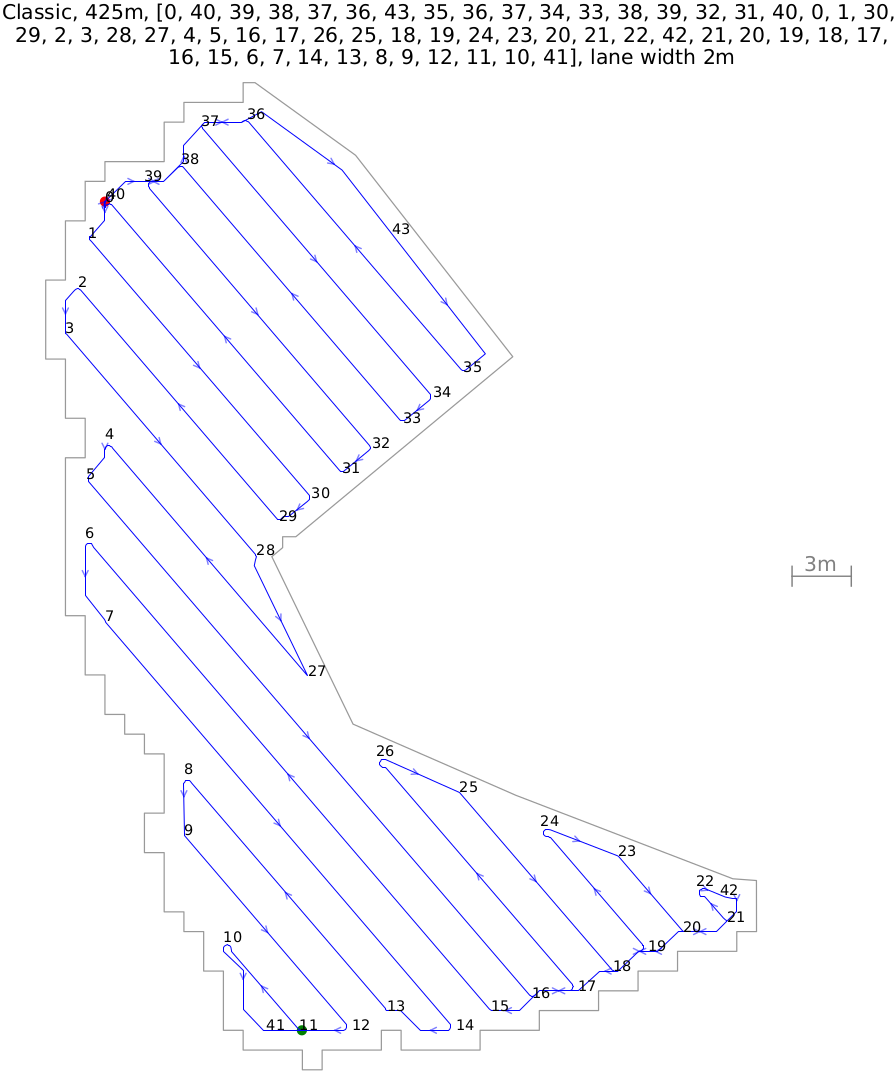}
\caption{Boustrophedon path for Fig. \ref{fig_Boustrophedon_areacovg}. Boustrophedon-pathlengths are much shorter since not including any headland coverage. However, on the flip-side they do not offer full area coverage. Since this is of predominant importance for agricultural applications, such as pesticide spraying and the like, Boustrophedon paths are not considered for the remainder of this paper.}
\label{fig_expt2_home}
\end{figure}

As discussed in Sect. \ref{sec_problmodeling} and visualised in Fig. \ref{fig_probvisualisation} there are patches of varying sizes. For $N_\text{patches,covg}\leq N_\text{patches,all}$ larger patches area coverage path planning is required to generate a structured path for the UAV to fly.

The by far pre-dominant area coverage path planning method for UAVs is based on \emph{Boustrophedon paths} (sketched in Fig. \ref{fig_Boustrophedon}), see e.g. \cite{xie2020path}, \cite{huang2023autonomous}, \cite{li2023coverage}, \cite{huang2023autonomous} and \cite{coombes2017boustrophedon}.

In this paper, however, it is argued that Boustrophedon-based coverage path planning is unsuitable for precision spot spraying applications.
Two reasons are given. First, characteristic for Boustrophedon-paths are area coverage \emph{gaps}, see Fig. \ref{fig_Boustrophedon_areacovg}. Second, one method to counteract these gaps is to enlarge obstacles area (e.g. in \cite{pham2020complete}) or, in spirit similar, artificially enlarge patches areas such that original core patches areas are fully covered. This method however fails when patches areas are directly bordering the field area contour (which is exactly the case for the experimental real-world data in this paper). Furthermore, this method of artificially enlarging patches areas then entails the danger of overspraying healthy field areas (i.e. beyond the original core patches areas).  

For these reasons, in this paper it is suggested for precision spot spraying to include a \emph{headland path} for area coverage path planning. Therefore, two methods are considered, which are sketched for a rectangular area in Fig. \ref{fig_sketch_2pathplanningmethods}:
\begin{itemize}
\item[(i)] A Boustrophedon-path extended with a headland path shall in the following be referred to as the \textsf{Classic} area coverage path planning method,  since it is the classic method used for ground-based area coverage in agricultural fields.
\item[(ii)] An optimised area coverage path planning method according to \cite{plessen2019optimal} is referred to as \textsf{Optimised} in the following, minimising area coverage pathlength under the assumption of existence of a headland path.
\end{itemize}

The benefit of including a headland path for area coverage path planning in comparison to Boustrophedon-based planning is visualised in Fig. \ref{fig_areacovg}.

A detail is discussed. In Fig. \ref{fig_incldheadl_areacovg} small area coverage gaps remain. The reason is a particularly edgy area contour that stems from the fact that area contour coordinates were here transfered from images captured by an UAV. One method to correct those area coverage gaps is to apply a \emph{smoothing} technique such as in \cite{plessen2024smoothing}. The simplest method, however, for this case of pixel-like area contours is to calculate the \emph{convex hull} area contour. (The convex hull area calculation is also used in the next subsection to discuss obstacle avoidance.) The disadvantage of any smoothing step is that original core patches areas are modified and potentially enlarged (such that additional healthy field area would be sprayed afterwards). For emphasis and illustration of this detail, the small area coverage gaps as Fig. \ref{fig_incldheadl_areacovg} are kept in this paper. The importance of input data will again be discussed in Section \ref{subsec_extendingdiscussion} when discussing a detected input data anomaly.

\subsection{Obstacle avoidance method\label{subsec_obstavoidmethod}}

\begin{figure}
\centering
\includegraphics[width=.35\linewidth]{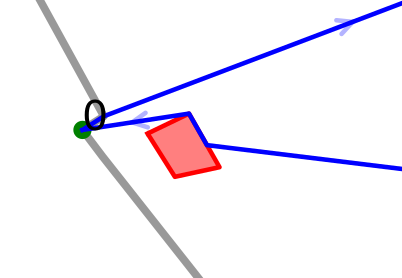}\hspace*{1.cm}
\includegraphics[width=.35\linewidth]{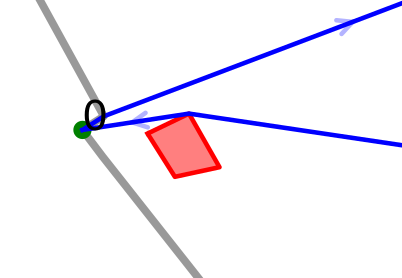}
\caption{Two obstacle avoidance methods: (left) approach the obstacle contour, follow its contour until a direct line-of-sight transition towards the next waypoint becomes feasible; (right) plan a global obstacle avoding path directly via a tangent path towards the obstacle and away from the obstacle.}
\label{fig_expt2_obstavoid}
\end{figure}

\begin{figure*}
\centering
\begin{subfigure}[t]{.33\textwidth}
  \centering
  \includegraphics[width=.99\linewidth]{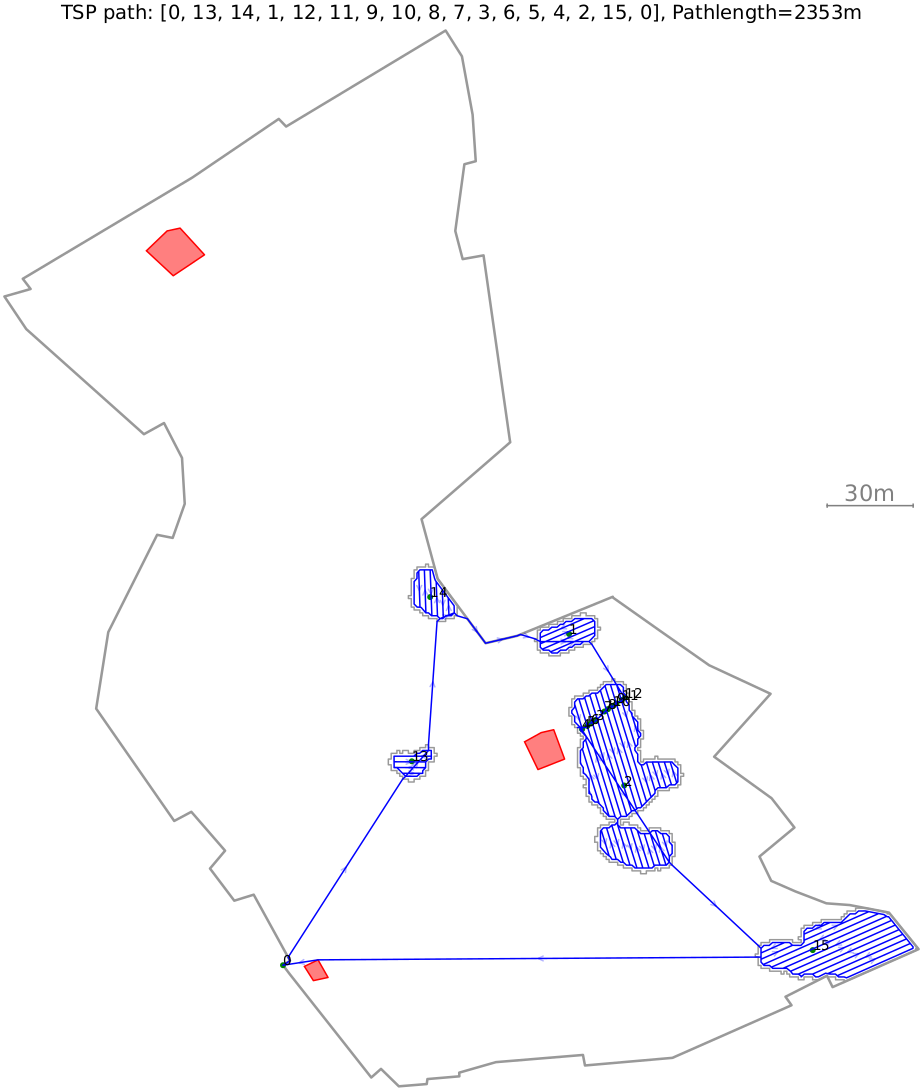}
\caption{Expt. 1: 15 patches}
  \label{fig_expt1}
\end{subfigure}%
\begin{subfigure}[t]{.33\textwidth}
  \centering
  \includegraphics[width=.99\linewidth]{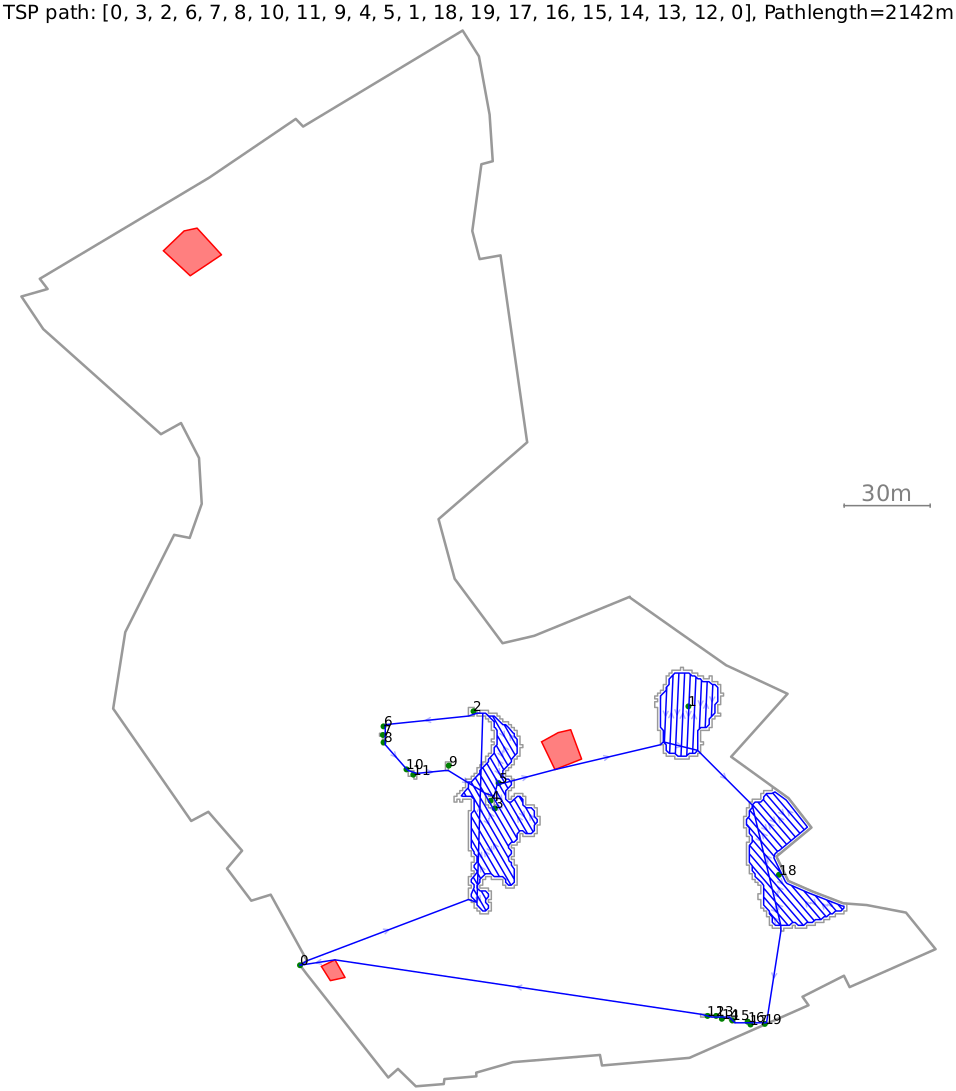}
\caption{Expt. 2: 19 patches}
  \label{fig_expt2}
\end{subfigure}
\begin{subfigure}[t]{.33\textwidth}
  \centering
  \includegraphics[width=.99\linewidth]{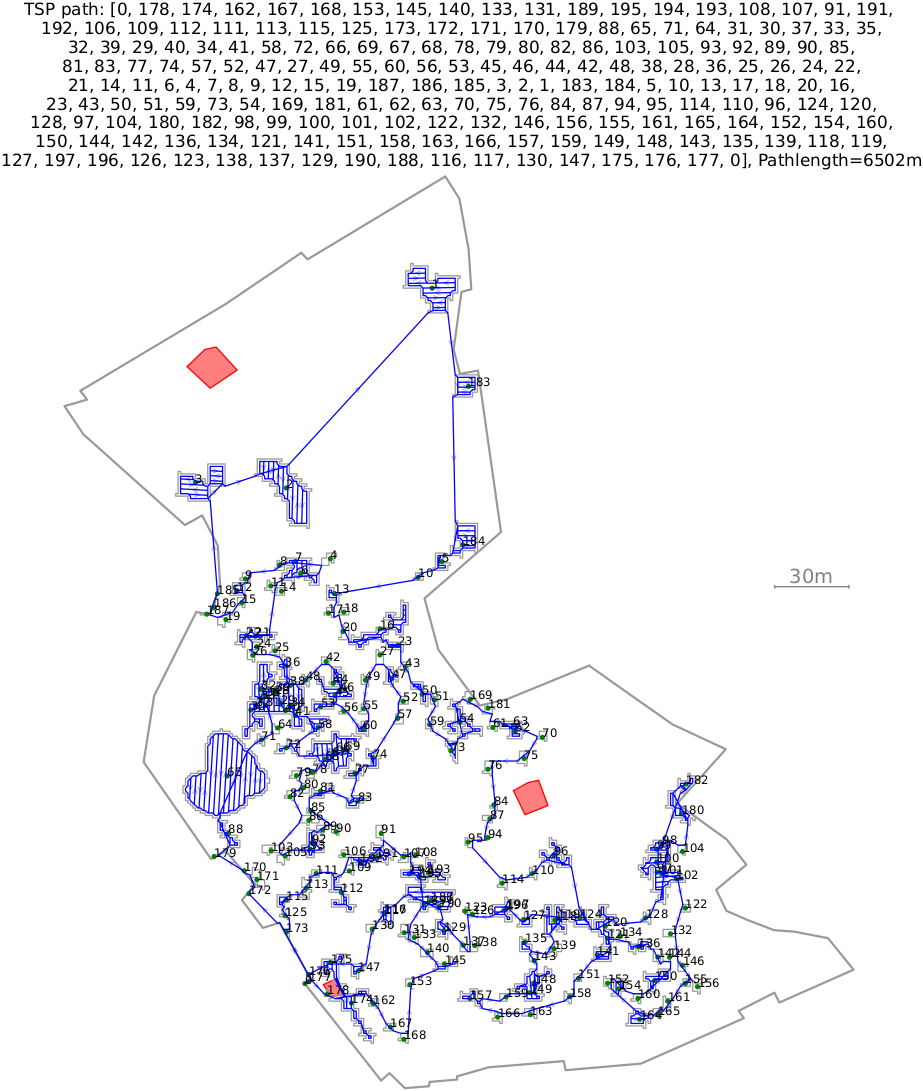}
\caption{Expt. 3: 197 patches}
  \label{fig_expt3}
\end{subfigure}
\caption{Results of 3 experiments. See also Tables \ref{tab1}-\ref{tab3}. For each experiment, the total path is composed of (i) a TSP-path connecting the patches (each patch is numbered, the TSP-path is given by the numbers sequence in the title of each figure), and (ii) area coverage paths optimised for each of the patches.}
\label{fig_3expts_results}
\end{figure*}

\begin{table}
\vspace{0.3cm}
\centering
\begin{tabular}{l|ll}
\hline \hline
Parameter & $T_\text{limit}$ & $W$ \\[1pt]
\hline
Value & 10s & 2m \\
\hline
\end{tabular}
\caption{Parameters used throughout all numerical experiments: a time limit for random sampling-based heuristics \textsf{H$_1$} to \textsf{H$_3$}, and the UAV operating width.}
\label{tab_param}
\end{table}

The work area might include obstacle areas such as ponds, which might pose a danger of irretrievable UAV-loss, or tree islands that shall be avoided by the planned path. A simple obstacle avoidance method is presented.

In general, obstacle areas can be arbitrarily non-convexly shaped. In a first step, for each obstacle its area contours are replaced by their \emph{convex hull} area contours. This is merited since by geometric arguments a shortest path around an obstacle always follows the convex hull area contours (rather than the non-convexly shaped contours). In a second step, it is differentiated between two obstacle avoidance methods that are visualised in Fig. \ref{fig_expt2_obstavoid}. They differ in how the obstacle is approached by a planned path. In one method, the UAV would fly direcly towards the obstacle, before following the obstacle area contour until the next waypoint (i.e. the next patch area) becomes visible. In another method, a global approach is taken that plans an obstacle avoding path directly as a tangent path towards the obstacle. The second part, i.e. the tangent path from the obstacle away and towards the next waypoint, is coinciding for both methods.

Notice that the proposed obstacle avoidance method can similarly be used to ensure planned paths stay within field area contours. In fact, this will be further discussed in Section \ref{subsec_extendingdiscussion}, when analysing a data anomaly for one of the experiments. 

Several comments are made. First, the latter method (global approach) offers a shorter overall pathlength and is used for the remainder of this paper. Second, this obstacle avoidance method can be generalized to landmark-based navigation (\cite{babel2014flight}) and further to the geometric corridor planner in \cite{plessen2017spatial}. Third, in practice obstacle areas might be artificially enlarged to generate a safety zone. Fourth, an implementational detail is discussed. Given input data is assumed such that obstacle areas are not overlapping with patches areas that are meant to be sprayed by the UAV. Therefore, obstacle area avoidance is only relevant for the TSP-path (and not applicable for the area coverage path planning part for the patches). Moreover, the TSP-path is generated by connecting the projection points and shortest distances between patches and thus involves much less waypoints than the area coverage paths. Thus, for obstacle area avoidance only the TSP-path has to be processed which entails processing of few waypoints, which is beneficial for efficiency.

\section{Numerical Results, Discussion and Limitations\label{sec_IllustrativeEx}}


\begin{table*}
\vspace{0.3cm}
\centering
\begin{tabular}{|l|rr|rr|rr|}
\hline \hline
 & \multicolumn{2}{c|}{Expt. 1} & \multicolumn{2}{c|}{Expt. 2} & \multicolumn{2}{c|}{Expt. 3} \\ \hline
Method & \multicolumn{1}{c}{$T$} & \multicolumn{1}{c|}{$L_\text{TSP}$} & \multicolumn{1}{c}{$T$} & \multicolumn{1}{c|}{$L_\text{TSP}$} & \multicolumn{1}{c}{$T$} & \multicolumn{1}{c|}{$L_\text{TSP}$} \\[1pt]
\hline
\textsf{NN} & \textbf{0.000035s} & \textbf{431.76m}  & \textbf{0.000046s}  & \textbf{416.52m} & 0.010483s & 1616.76m\\
\textsf{DENN} & 0.000047s & 486.50m  & 0.000060s  & 508.86m & \textbf{0.011337s} & \textbf{1574.04m}\\
\hline
\end{tabular}
\caption{Quantitative Results 1/3: Evaluation of 2 TSP-initialisation methods.}
\label{tab1}
\end{table*}

\begin{table*}
\vspace{0.3cm}
\centering
\begin{tabular}{|c|rr|rr|rr|}
\hline \hline
 & \multicolumn{2}{c|}{Expt. 1} & \multicolumn{2}{c|}{Expt. 2} & \multicolumn{2}{c|}{Expt. 3} \\[-4pt] 
  & \multicolumn{2}{c|}{\footnotesize{(start from NN)}} & \multicolumn{2}{c|}{\footnotesize{(start from NN)}} & \multicolumn{2}{c|}{\footnotesize{(start from DENN)}} \\ \hline
Heuristics & \multicolumn{1}{c}{$T$} & \multicolumn{1}{c|}{$L_\text{TSP}$} & \multicolumn{1}{c}{$T$} & \multicolumn{1}{c|}{$L_\text{TSP}$} & \multicolumn{1}{c}{$T$} & \multicolumn{1}{c|}{$L_\text{TSP}$} \\[1pt]
\hline
only \textsf{H$_1$} & 10s & 388.03m  & 10s  & 416.52m & 10s & 1399.60m\\
only \textsf{H$_2$} & 10s & 388.03m  & 10s  & 413.82m & 10s & 1205.76m\\
only \textsf{H$_3$} & 10s & 431.76m  & 10s  & 416.52m & 10s & 1532.77m\\
only \textsf{H$_4$} & \textbf{0.00019336s}  & \textbf{383.16m}  & 0.00061631s & 403.19m & 0.22768021s & 1119.76m\\
\{\textsf{H$_2$},\textsf{H$_4$}\} & 10.00020337s & 383.16m  & \textbf{10.00036430s}  & \textbf{401.97m} & \textbf{10.25113702s} & \textbf{1069.55m}\\
\{\textsf{H$_1$},\textsf{H$_2$},\textsf{H$_3$},\textsf{H$_4$}\}  & 30.00021315s & 383.16m  & 30.00037289  & 401.97m & 30.14522696s & 1070.63m\\
\hline
\end{tabular}
\caption{Quantitative Results 2/3: Evaluation of 4 TSP-refinement heuristics plus 2 additional combinations. For each experiment it is started from the best initialisation (\textsf{NN} or \textsf{DENN}) according to Table \ref{tab1}.}
\label{tab2}
\end{table*}

\begin{table}
\vspace{0.3cm}
\centering
\begin{tabular}{|l|r|r|r|}
\hline \hline
 & \multicolumn{1}{c|}{Expt. 1} & \multicolumn{1}{c|}{Expt. 2} & \multicolumn{1}{c|}{Expt. 3} \\[2pt] \hline
\scriptsize{{\color{gray}\%Nr. of larger patches w/ coverage paths}} &  &   &   \\ 
$N_\text{patches,all}$ & 15 & 19  & 197  \\ 
$N_\text{patches,covg}$ & 5 & 3  & 58  \\[2pt] \hline
\scriptsize{{\color{gray}\%Coverage pathlengths for two methods}} &  &   &   \\ 
$\sum_{i=1}^{N_\text{patches,covg}} L_i^\text{classic}$ & 2086m & 1848m  & 5178m  \\
$\sum_{i=1}^{N_\text{patches,covg}} L_i^\text{optim}$ & 1934m & 1698m  & 4740m  \\
Savings [m] & -152m & -150m  & -438m \\
Savings [\%] & \textbf{-7.3\%} & \textbf{-8.1\%}  & \textbf{-8.5\%} \\[2pt] \hline
\scriptsize{{\color{gray}\%Total pathl. (TSP plus coverage paths)}} &  &   &   \\ 
$ L_\text{total}^\text{w\textbackslash classic}$ & 2505m & 2292m  & 6940m \\
$ L_\text{total}^\text{w\textbackslash optim}$ & 2353m & 2142m & 6502m \\
Savings [m] & -152m & -150m  & -438m \\
Savings [\%] & \textbf{-6.1\%} & \textbf{-6.5\%}  & \textbf{-6.3\%} \\[2pt] \hline
\scriptsize{{\color{gray}\%Percentg. coverage pathl. of total pathl.}} &  &   &   \\ 
$ (\sum_{i=1}^{N_\text{patches,covg}} L_i^\text{classic}) / L_\text{total}^\text{w\textbackslash classic}$ & 83.3\% & 80.6\%  & 74.6\%  \\
$ (\sum_{i=1}^{N_\text{patches,covg}} L_i^\text{optim}) / L_\text{total}^\text{w\textbackslash optim}$ & \textbf{82.2\%} & \textbf{79.3\%}  & \textbf{72.9\%}  \\
\hline
\end{tabular}
\caption{Quantitative Results 3/3: Evaluation of the contributions of TSP-related and area coverage-related pathlenghts on total pathlengths, respectively.}
\label{tab3}
\end{table}

\begin{figure*}
\centering
\begin{subfigure}[t]{.49\textwidth}
  \centering
  \includegraphics[width=.93\linewidth]{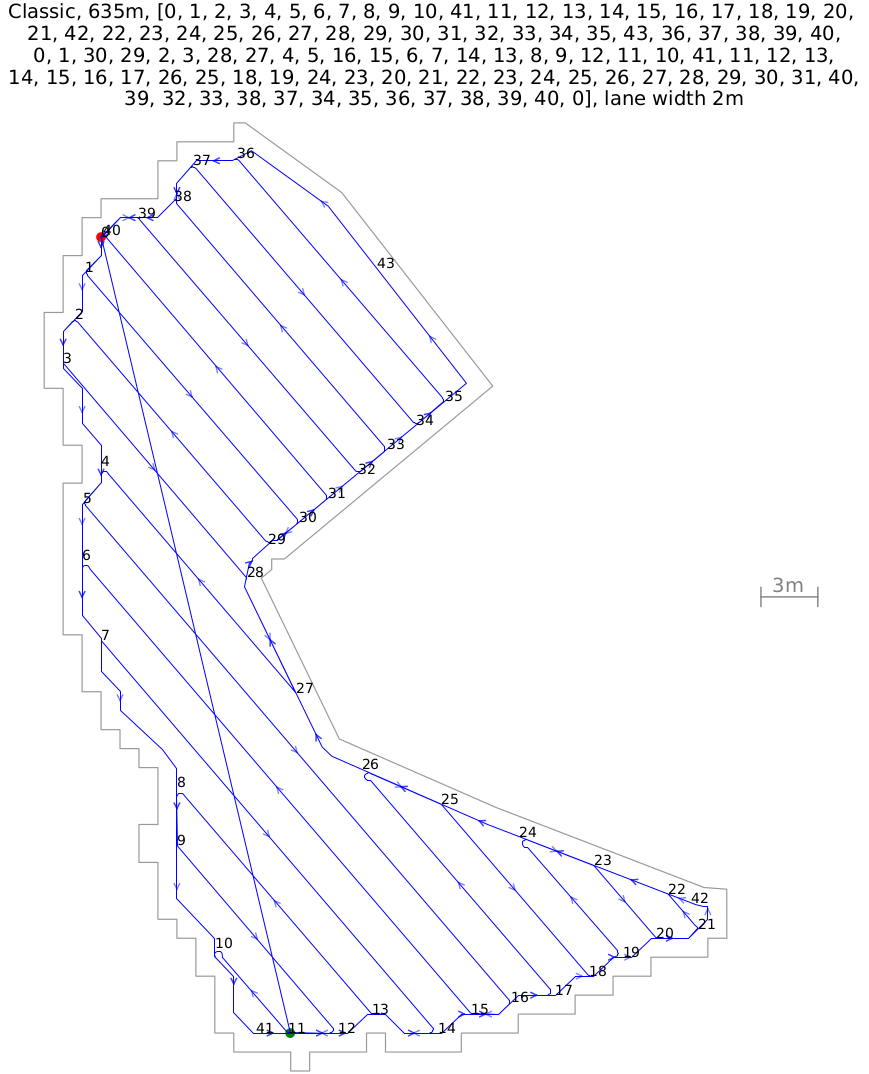}
\caption{Classic: pathlength 635m}
  \label{fig_classic}
\end{subfigure}
\begin{subfigure}[t]{.49\textwidth}
  \centering
  \includegraphics[width=.93\linewidth]{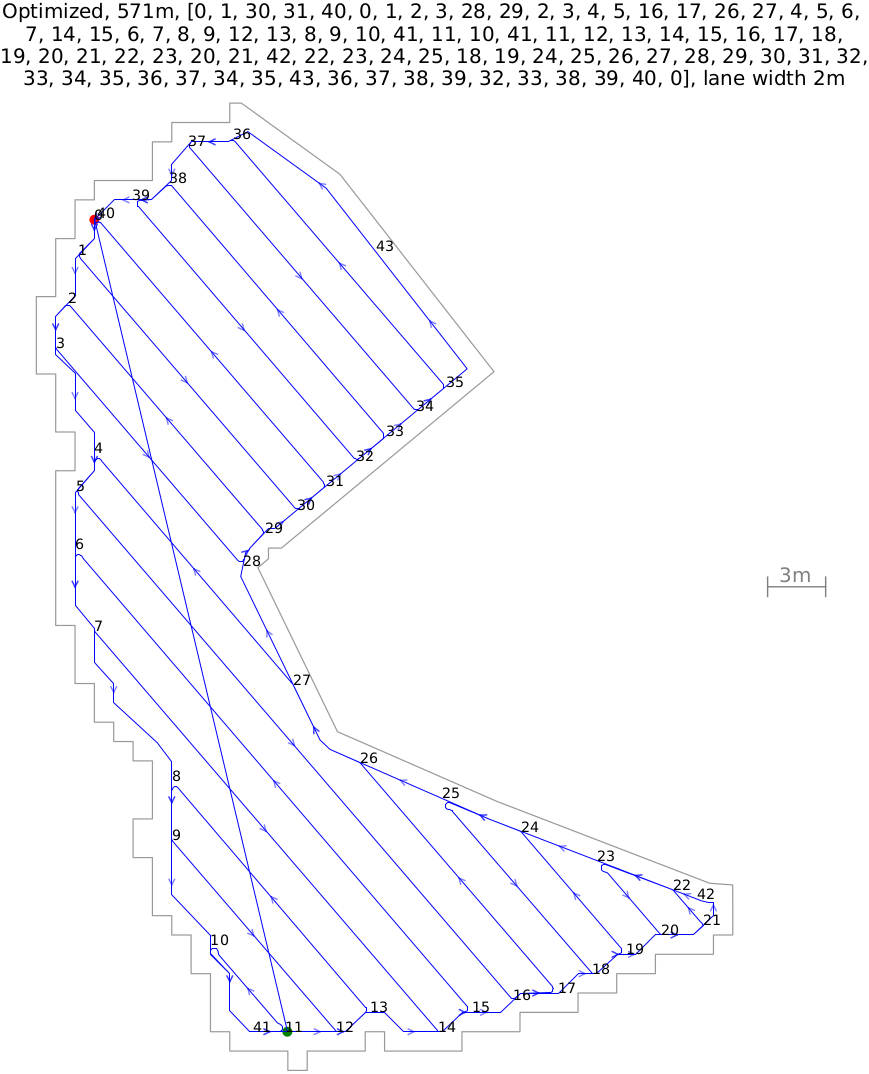}
\caption{Optimised: pathlength 571m}
  \label{fig_optim}
\end{subfigure}
\caption{Two  path planning methods for area coverage (\cite{plessen2019optimal}). Starting at node 0 a path is computed (node sequences in the title starting and ending at node 0). Afterwards, the UAV flies from node 0 to the projection point (green dot at the plot bottom) that connects towards the next patch (see also Fig. \ref{fig_patchentryexit}).}
\label{fig_areacovg_classicoptim}
\end{figure*}

Proposed methods are evaluated on a real-world field of size 5.2ha with three obstacle areas and three different experimental settings with 15, 19 and 197 patches, respectively. For final path planning results see Fig. \ref{fig_3expts_results}. Table \ref{tab_param} summarizes parameters. A solve time limit of $T_\text{limit}=10$s is imposed for the random sampling-based TSP-refinement heuristics. An UAV operating width of $W=2$m is assumed. All experiments were run on a laptop running Ubuntu 22.04 equipped with an Intel Core i9 CPU @5.50GHz×32 and 32 GB of memory. The evaluation of different TSP-path initialisation and refinement heuristics as well as the two area coverage path planning methods is summarised in Tables \ref{tab1}-\ref{tab3} and discussed in detail in the following.

First, as Table \ref{tab1} shows in two of the three experiments NN outperforms DENN. However, in the large-scale experiment with 197 patches DENN outperforms. In all experiments NN has a slightly smaller solve time.

Second, while determining the best TSP-initialisation scheme was inconclusive this is radically different for the TSP-refinement step. Here, heuristic \textsf{H$_4$} consistently performed best among the 4 heuristics and results in minimal solve times due to its deterministic nature. \textsf{H$_1$} to \textsf{H$_3$} are random sampling-based methods such that the solve time limit is always reached. The suitability of heuristic \textsf{H$_4$} for the problem at hand is the first key finding of this paper. When combining heuristics \textsf{H$_2$} and \textsf{H$_4$} the lowest-cost TSP-solutions could be determined for Experiments 2 and 3, however at the cost of increased solve times in comparison to when employing only \textsf{H$_4$}.

Third, Table \ref{tab3} evaluates (i) the two area coverage path planning methods \textsf{Classic} and \textsf{Optimised}, and (ii) evaluates the contribution of area coverage pathlengths on total pathlength in comparison to TSP-related pathlengths. The \textsf{Optimised}-method can save between -7.3\% to -8.5\% in comparison to the \textsf{Classic}-method. Including the TSP-related pathlength, which is the same for both methods, these savings are diluted to -6.1\% to -6.3\% for total pathlengths. Crucially and surprisingly, the contribution of the sum of all area coverage pathlengths on the total pathlength is very high with 72.9\% to 82.2\%. This is a second key finding of this paper. Note that this high percentual contribution is despite the fact that only 5, 3 and 58 patches of the total 15, 19 and 197 patches are large enough to require area coverage planning. There are three reasons: (i) inclusion of headland path for area coverage planning as motivated in Sect. \ref{subsec_patchesareacovg}, (ii) the inclusion of the pathlength of the transition of the patch entry to the patch exit included in the contribution, and (iii) the existence of large patches areas as shown in Fig. \ref{fig_3expts_results}. The implication of this finding is that for spot spraying applications with UAVs hightened attention has to be attributed to area coverage path planning for the patches.

Fourth, Fig. \ref{fig_areacovg_classicoptim} visualises area coverage path planning for the two methods \textsf{Classic} and \textsf{Optimised} in detail. Notice that patches areas are in general arbitrarily non-convexly shaped, therefore necessitating a mathematical logic to determine optimised paths in a structured manner.

\subsection{Extending Discussion and Limitations \label{subsec_extendingdiscussion}}

\begin{figure}
\centering
\includegraphics[width=.3\linewidth]{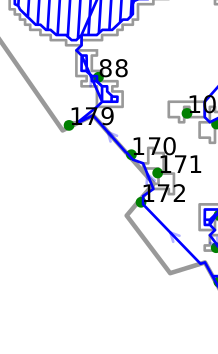}\hspace*{1.5cm}
\includegraphics[width=.3\linewidth]{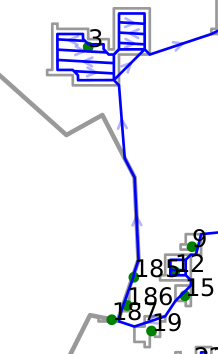}
\caption{Extending discussion 1/3. After manual correction of given 'faulty' data (coordinates for two patches were slightly exceeding field contours) at two locations, the resulting UAV-paths now stay within field contours. See the discussion in Sect. \ref{subsec_extendingdiscussion}.}
\label{fig_expt3_datacorrect}
\end{figure}

\begin{figure}
\centering
\includegraphics[width=.5\linewidth]{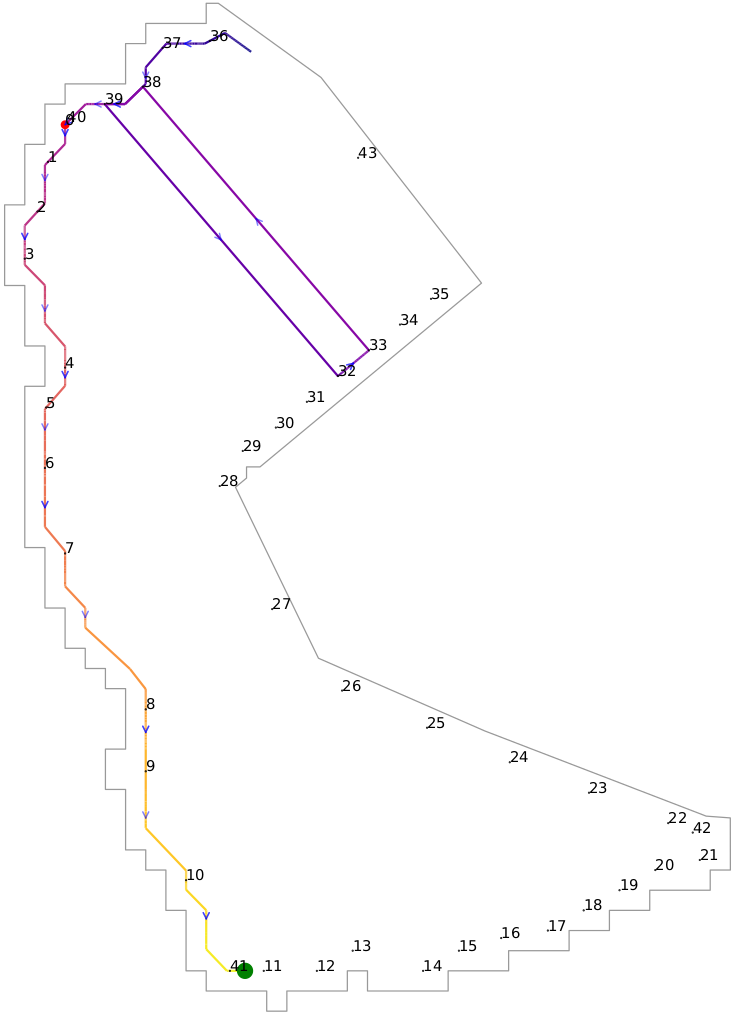}~
\includegraphics[width=.5\linewidth]{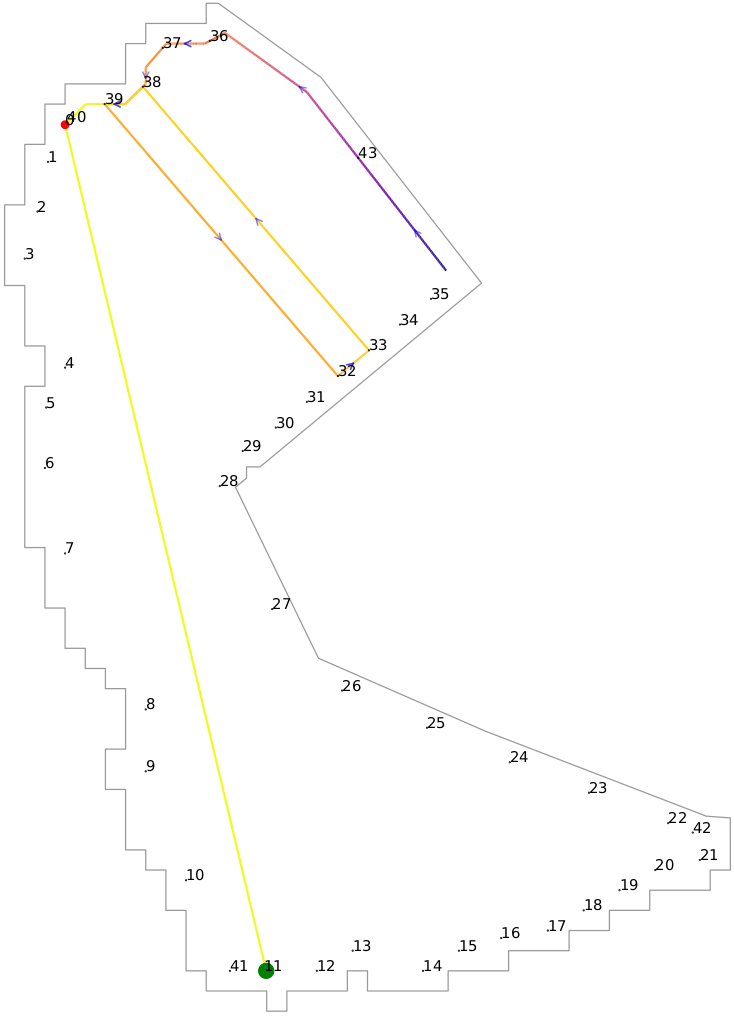}
\caption{Extending discussion 2/3. After area coverage two options are considered for the transition towards the patch exit point (green dot in the figures) that connect the path to the next patch. (Left) Transition along the headland path, (Right) Direct line-of-sight transition. The paths in the two subfigures are colored to indicate transition direction from darker color towards brighter color.}
\label{fig_expt2_home}
\end{figure}

\begin{figure}
\centering
\includegraphics[width=.5\linewidth]{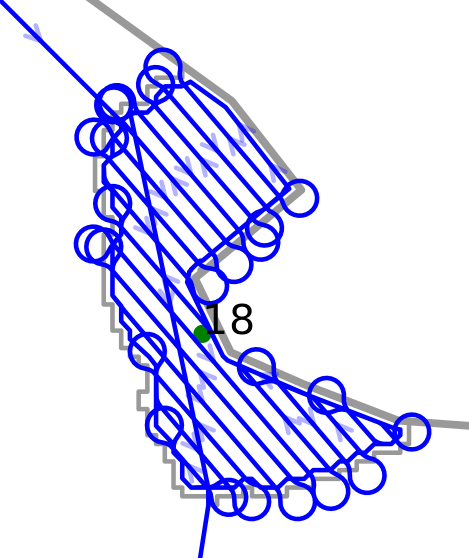}
\caption{Extending discussion 3/3. Path planning with limited turning radius, relevant especially for fixed-wing UAV.}
\label{fig_bulbs}
\end{figure}

This subsection discusses details, extending variations and limitations. First, when zooming into Fig. \ref{fig_expt3} one can observe two locations where planned paths unexpectedly exceed field area contours (in the vicinity of patches numbered 179 and 185). When further analysing this it was found that the reason was faulty input data (data was used 'as is'), where coordinates for the two patches were slightly exceeding field contours and the calculated projection points connecting patches in the vicinity were consequently located slightly \emph{outside} the field area. After manual correction of the data (by slightly modifying field contour coordinates at the two locations), the resulting UAV-path stays within field contours as Fig. \ref{fig_expt3_datacorrect} illustrates (thereby also validating the method from Sect. \ref{subsec_obstavoidmethod} used to correct TSP-paths to stay within field area contours). This anomaly detection example underlines the importance of input data verification and potential dangerous impacts (UAV-paths exceeding the field area contour) inaccurate input data may have.

Second, Fig. \ref{fig_expt2_home} visualises an alternative method for the transition to the patch exit point after area coverage (recall Fig. \ref{fig_patchentryexit}). Instead of a straight line a path tracing the headland may be used. This increases pathlength but may be of interest in scenarios where unneeded patch traversals shall be avoided (e.g. due to a leaking spray tank).

Third, a perceived strength of proposed method is that only two parameters are needed, see Table \ref{tab_param}. This is complexified when accounting for more detailed UAV-dynamics such as a specific turning radius or other dynamical vehicle constraints. Here, the method from \cite{plessen2024smoothing} (originally devised for ground-based motion) can be employed before adding a z-coordinate reference, either (i) by a constant offset distance tracking the field topograhpy, or (ii) by a piecewise-affine reference as suggested in \cite{lin2014path}, where a 2D Dubins curve is first created before a z-coordinate reference is added. 

An important aspect with respect to z-coordinate and 3D field topography is made. The method presented in this paper is a graph optimisation method, based on a transition graph model according to Section \ref{sec_problmodeling}. Therefore, on the graph optimisation level the presented method does not change for both 2D- and 3D-path planning. What changes, however, in those two cases is the \emph{mapping} from graph model to actual path coordinates, which then are either 2D or 3D coordinates.

The inclusion of UAV-dynamics with limited turning radii, which is especially relevant for fixed-wing aircraft and useful for larger patches areas, renders path planning more complex. See Fig. \ref{fig_bulbs} for exemplary illustration of a path plan when the UAV-turning radius is larger than half the operating width. This figure illustrates the benefits of quadrotor-like UAVs with near-holonomic flight dynamics that can avoid large turning radii characteristic for fixed-wing UAVs. Furthermore, flight maneuvers \emph{exceeding} the patch area as shown in Fig. \ref{fig_bulbs} offer the possibility of counteracting the downsides of Boustrophedon paths by enabling full area coverage by shifting the turning outside of the patch area. The flip-side, however, is that such maneuvers then might violate (i.e. exceed) the field contour, especially when a patch area is located close to the field contour (as is precisely the case for this particular patch, see patch 18 in Fig. \ref{fig_expt2}). 3D precision trajectory generation for UAVs is a topic of future work, also involving the planning of multiple cascaded headland paths such that flight maneuevers are contained to the patches areas even for turning radii larger than half the UAV operating width. Quadrotors offer the benefit of near-holonomic maneuverability. But fixed-wing UAVs might be more suited for large-scale areas.

A limitation of this paper is the absence of´ real-world flight experiments. The scope of this paper is nominal mission planning, in particular, path \emph{planning} (not path tracking) according to Problem \ref{problem1}, for which a high-level algorithm according to Fig. \ref{fig_blockdiag} is presented. The method is evaluated on real-world data.  

Finally, for the experiment of Fig. \ref{fig_3expts_results} the fitting of straight lanes to the patches areas was assumed. However, likewise the method from \cite{plessen2021freeform} can directly be applied to fit \emph{freeform} lanes (typically aligned to a patches area contour segment) to minimise the number of transitions between lanes and headland path.

\section{Conclusion\label{sec_conclusion}}

A path planning method for spots spraying with UAVs was presented, integrating a traveling salesman problem (TSP) solution with area coverage path planning. Assumptions were provided that permit to calculate suitable TSP-paths separately from the area coverage planning for the patches areas without loss of optimality. For the TSP-part a variety of heuristics was discussed, whereby one refinement heuristic encouraging the avoidance of edgy TSP-paths was found to be particularly useful. For the area coverage path planning, the inclusion of a headland path for area coverage gap avoidance was discussed. The surprising high contribution of area coverage pathlengths on the total pathlength (TSP-solution plus area coverage pathlengths for larger patches areas) was emphasised.  

In future work, the findings about TSP-heuristics will be transfered to multi-vehicle routing problems (VRPs) such as in \cite{plessen2019coupling}, where it was found difficult to scale exact integer linear program formulations to larger dimensions. 

In this perspective an interesting avenue of future work is multi-UAV path planning. Two options are envisioned. First, UAVs fly in a swarm and in parallel for patches areas coverage, which as a consequence permits to assume larger operating widths. Second, UAVs fly separately such that each UAV separetely treats a subset of the given patches. In this scenario the original operating width is maintained.

\bibliographystyle{model5-names} 
\bibliography{mybibfile.bib}
\nocite{*}







\end{document}